\documentclass[conference]{IEEEtran}
\IEEEoverridecommandlockouts
\usepackage{cite}
\usepackage{amsmath,amssymb,amsfonts}
\usepackage{algorithmic}
\usepackage{graphicx}
\usepackage{textcomp}
\usepackage{xcolor}
\usepackage{float}
\usepackage{tabularx}
\usepackage{booktabs} 
\usepackage{multirow}
\usepackage{pifont}
\usepackage[most]{tcolorbox}
\def\BibTeX{{\rm B\kern-.05em{\sc i\kern-.025em b}\kern-.08em
    T\kern-.1667em\lower.7ex\hbox{E}\kern-.125emX}}
\begin{document}
\bstctlcite{IEEEexample:BSTcontrol}
\title{Agentic IoT: Architectures, Applications, and Challenges Toward the Internet of Agents\\
%Agentic IoT: Toward Autonomous and Self-Adaptive Systems in the Internet of Agents\\
%\thanks{Identify applicable funding agency here. If none, delete this.}
}

\author{Rümeysa Hilal Sevinç, Bahaeddin Türkoğlu, and İbrahim Kök%
\thanks{The authors are with the Department of Artificial Intelligence and Data Engineering, Ankara University, Ankara, Türkiye.}%
\thanks{Corresponding author: İbrahim Kök (e-mail: ikok@ankara.edu.tr).}%
}

\maketitle

\begin{abstract}
The integration of AI into Internet of Things (AIoT) systems has gradually transformed them from passive data collection infrastructures into intelligent systems capable of anomaly detection, predictive maintenance, classification, forecasting, and optimization. However, most existing solutions still rely on task-specific models that infer from sensor data; thus, system-wide capabilities such as real-time reasoning, adaptive planning, autonomous coordination, learning, tool use, and contextual decision-making remain limited. This paper examines Agentic IoT as a next-generation cognitive IoT paradigm that integrates the perception, reasoning, planning, learning, and action capabilities of autonomous AI agents with cyber-physical systems. Agentic IoT aims to transform IoT from data-centric sensing and inference infrastructures into distributed cognitive agent ecosystems operating across the device/edge–fog–cloud continuum. The paper first grounds this transition as a paradigm shift and positions Agentic IoT in relation to AIoT, edge intelligence, multi-agent systems, and the Internet of Agents. It then systematically reviews current studies, presents a holistic architectural framework, discusses domain-specific application potential, and identifies key technical, operational, and research challenges together with future research directions.

\end{abstract}

\begin{IEEEkeywords}
Agentic IoT, Internet of Agents, Internet of Things, Autonomous Agents, Large Language Models(LLMs)
\end{IEEEkeywords}

\section{Introduction}

In recent years, rapid developments in Large Language Models (LLMs), autonomous AI agents, and the design and development of intelligent systems have initiated a fundamental transformation \cite{11202920}. While traditional machine and deep learning models effectively perform classification, prediction, or optimization tasks, today's agentic AI systems are capable of perceiving contextual information, performing reasoning, creating multi-step plans, utilizing external tools, learning from feedback, and collaborating with other agents to accomplish complex goals \cite{abou2025agentic}. Thanks to these advanced capabilities, a significant transformation is currently taking place from reactive AI systems toward a world of autonomous and goal-oriented intelligent agents.

The implications of this transformation are particularly significant for the Internet of Things (IoT). As the number of heterogeneous devices connected to IoT increases, the management of heterogeneous devices, dynamic operating conditions, and real-time decision-making requirements have also increased exponentially \cite{kok2019deep}. Smart-X approaches in many domains, such as smart homes, energy smart environments, smart cities, intelligent transportation networks, and precision agriculture applications, increasingly highlight the requirements for distributed decision-making, contextual awareness, real-time coordination, autonomous operation, and adaptation to changing environmental conditions \cite{Dabels2023SmartX}. Therefore, next-generation IoT systems are expected not only to collect data and provide communication, but also to understand context, make autonomous decisions, coordinate their actions, and adapt to changing environmental conditions.

In line with these needs in the IoT domain, as in many other fields, new concepts have emerged aiming to enable autonomous agents to operate effectively in large-scale and distributed systems \cite{10849561}. For example, recent developments in agentic AI have inspired the emergence of a broader vision referred to as the Internet of Agents (IoA) \cite{IoA_survey}. IoA evolves the traditional Internet paradigm from human-centered communication toward large-scale interaction, coordination, and collaboration among autonomous intelligent agents. The IoA approach places autonomous agents at the center of decision-making and computation processes. Within this vision, intelligent agents are envisioned to discover services, negotiate tasks, share knowledge, coordinate actions, and collectively solve complex problems \cite{dazzi2025internet}.

Integrating the conceptual and methodological vision provided by IoA into IoT ecosystems involves unique opportunities and challenges. IoT environments possess characteristics such as resource constraints, distributed infrastructures, heterogeneous devices, intermittent connectivity, real-time requirements, and direct interaction with the physical world \cite{9099866}. These characteristics require a specialized perspective that integrates the capabilities of autonomous agents with the operational realities of IoT systems. In this direction, we propose the concept of Agentic IoT, which refers to the integration of autonomous agent capabilities with IoT systems, and position it as an application-oriented extension of the Internet of Agents vision in the IoT domain.

Within the scope of Agentic IoT, IoT systems evolve beyond mere data collection and monitoring infrastructures into cognitive ecosystems capable of making proactive decisions, adapting to changing environmental conditions, accomplishing goal-oriented tasks, and collaborating with other agents \cite{11184150}. At this point, the task-oriented operating model of AIoT approaches gives way to autonomous, context-aware, and action-oriented systems.
\begin{table*}[t]
\centering
\caption{Comparison of IoT, AIoT, and Agentic IoT Paradigms}
\label{tab:comparison}
\renewcommand{\arraystretch}{1.35}
\begin{tabular}{p{3.3cm} p{3.3cm} p{3.9cm} p{5cm}}
\hline
\textbf{Feature} & \textbf{IoT} & \textbf{AIoT} & \textbf{Agentic IoT} \\
\hline
Decision Mechanism      & Rule-based/event driven              & Model-based (ML/DL)         & LLM-driven reasoning \\
Autonomy Level          & Low                    & Low/Medium                         & Medium/High \\
Reasoning Capability    & \ding{55}               & Task-spesific                   & Explicit reasoning (CoT, ReAct) \\
Planning                & Predefined workflows               & Optimization or RL-based                   & Task decomposition, dynamic replanning \\
Tool Use / API Calling  & System-level API integration               & Model-assisted service integration                   & Agentic tool use, Function calling \\
Memory                  & Logs, historical records               & Historical data, Model states                   & Short-term + Long-term \\
Learning                & \ding{55}               & Offline/online training, RL            & Continuous, Retrieval-augmented learning \\
Adaptivity              & Static                  & Limited/partially                     & Self-adaptive, Context aware \\
Multi-Agent Coordination & \ding{55}              & \ding{55}                   & Multi-agent collaboration (MCP, A2A) \\
Natural Language Interaction & \ding{55}           & \ding{55}                   & \ding{51} \\
Behavior Model          & Reactive                & Predictive                  & Proactive + Goal-driven \\
\hline
\end{tabular}
\end{table*}
Although some studies have emerged in the literature on the Internet of Agents \cite{wang2025internet}, Agentic AI \cite{sapkota2025ai}, multi-agent systems, and autonomous cyber-physical systems \cite{auto_cps,siakas2025autonomous}, to the best of our knowledge, there is still no comprehensive survey study that integrates these research domains from an IoT perspective and examines the concept of Agentic IoT in all its aspects. Although each of these research areas contributes to the development of autonomous IoT ecosystems, they often employ different terminologies, assumptions, architectural abstractions, and evaluation methodologies. As a result, the concept of Agentic IoT has not yet been clearly defined as a holistic and systematic research paradigm, and its relationship with related concepts has not been sufficiently clarified.

Motivated by the aforementioned research gaps, this study aims to systematically address the concept of Agentic IoT and establish a holistic foundation for this emerging research field. The main contributions of this study are summarized as follows:
\begin{itemize}
    \item We formally define the concept of Agentic IoT and explain its relationship with IoA, AIoT, and Agentic AI.
    \item We present the paradigm shift from traditional IoT and AIoT systems toward autonomous, agent-centric, and cognitive IoT ecosystems.
    \item We comprehensively review the current literature, enabling technologies, reference architectures, communication mechanisms, and prominent application domains of Agentic IoT.
    \item We identify the major challenges and future research directions related to the real-world deployment of Agentic IoT systems.

\end{itemize}
The remainder of this paper is organized as follows. Section \ref{sec_II} discusses the evolution from traditional IoT systems to Agentic IoT and the fundamental paradigm shifts underlying this transformation. Section \ref{sec_III}  presents the proposed Agentic IoT architecture, including the reference architecture, agent cognitive loop, memory and tool integration mechanisms, communication frameworks and the deployment strategies. Section \ref{sec_IV}  examines Agentic IoT applications across different domains. Section \ref{sec_V}  discusses the key challenges and outlines future research directions. Finally, Section VI concludes the paper.

\section{From Traditional IoT to Agentic IoT: A Paradigm Shift}
\label{sec_II}
In recent years, due to rapid and emerging developments in the field of AI, significant paradigm shifts have been experienced in the Internet of Things domain, as in many other fields and technologies. Traditional IoT systems were largely focused on collecting data from the physical environment and processing this data through event-driven mechanisms on cloud-centric infrastructures. In the following stage, computational capabilities were moved toward fog and edge networks, enabling real-time analytics and low-latency decision-making mechanisms \cite{koek2022content}. However, existing fog and edge computing approaches still exhibit reactive behavioral models running on static learning pipelines that are unable to inherently reason about environmental uncertainties, long-term objectives, causal relationships, and dynamic operational constraints. During this period, the concept of AIoT (Artificial Intelligence of Things) emerged through the integration of ML and AI algorithms into IoT systems to address existing challenges \cite{siam2025artificial}. Although AIoT is still used in many domains and systems, the models employed in these systems inherently operate on pre-trained static models. Therefore, they remain inadequate in situations that require heterogeneous, dynamic, uncertain, and real-time reasoning capabilities. These limitations have clearly revealed the need for a paradigm in IoT systems that can provide contextual understanding, autonomous decision-making, and adaptive behavior \cite{11184150}.
\begin{tcolorbox}[
colback=gray!5,
colframe=black,
title=\textbf{Definition: Agentic IoT},
fonttitle=\bfseries
\label={def:agentic-iot}
]
\textit{Agentic IoT} is a next-generation cognitive IoT paradigm that transforms IoT from networks of connected devices into distributed cognitive agent ecosystems by integrating the perception, reasoning, planning, learning, and action capabilities of autonomous AI agents with cyber-physical systems.
\end{tcolorbox}
\label{sec_III}
\begin{figure*}[!t]
    \centering
\includegraphics[width=\textwidth]{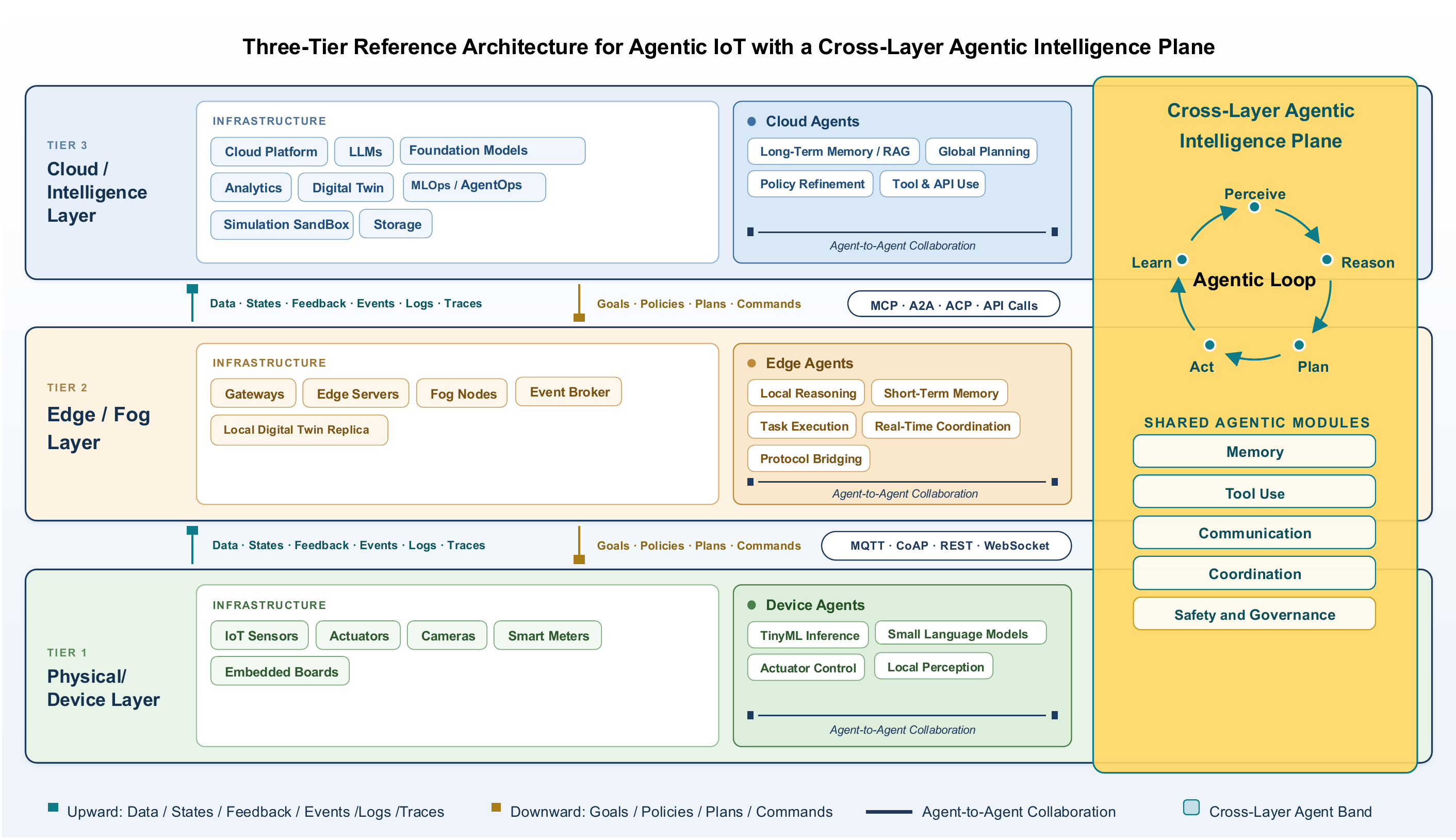}
    \caption{Three-Tier Reference Architecture for Agentic IoT}
    \label{Aiot_architecture}
\end{figure*}
With the development of LLMs, AI systems have gained advanced capabilities such as reasoning, planning, tool use, multi-step decision-making, and the interpretation of environmental context~\cite{11540994}. This evolution represents a paradigm shift in IoT research, where connected devices are no longer viewed merely as passive sensing and actuation endpoints, but as components of autonomous, context-aware, and goal-directed cyber-physical ecosystems. In this study, we use the term \textit{Agentic IoT} to denote this emerging paradigm, as formalized in the definition box.

\section{Agentic IoT Architecture}
\label{sec_III}
This section describes how the Agentic IoT vision can be realized as a concrete system design. The guiding principle is that agent intelligence should be distributed across the IoT stack in proportion to the resources available at each layer, while all agents share a common cognitive model and a common set of support modules.

\subsection{Reference Architecture}
Fig.~\ref{Aiot_architecture} illustrates the proposed three-tier reference architecture for Agentic IoT. The architecture organizes the system into a physical/device layer, an edge/fog layer, and a cloud/intelligence layer, and complements them with a cross-layer agentic intelligence plane that spans all three. Each tier combines conventional IoT infrastructure with a class of agents whose capabilities are matched to the computational resources available at that level.

The physical/device layer (Tier 1) contains the components that interact directly with the environment: IoT sensors, actuators, cameras, smart meters, and embedded boards. Device agents at this level operate under strict memory, computation, and energy budgets. They therefore rely on TinyML inference and small language models (Table~\ref{tab:iot_llm_comparison}) for local perception, and on actuator control modules to turn decisions into physical actions. Their role is deliberately narrow: fast perception--action behavior at the point where data is produced, without waiting for the upper layers.

The edge/fog layer (Tier 2) consists of gateways, edge servers, and fog nodes, supported by an event broker and a local replica of the digital twin. Edge agents reason over data aggregated from many devices, maintain short-term memory of the current operational context, execute tasks, and coordinate devices in real time. This tier also hosts protocol bridging, the mechanism that connects device-level IoT protocols with agent-level communication (Section~\ref{sec_III}D). Since edge agents sit close to the field but command more resources than any single device, they are the natural place for time-critical decisions that involve multiple devices.

The cloud/intelligence layer (Tier 3) provides the heaviest computational resources: cloud platforms, LLMs and foundation models, storage, analytics, digital twins, MLOps/AgentOps pipelines, and a simulation sandbox. Cloud agents use this infrastructure for the capabilities that lower tiers cannot afford: long-term memory with retrieval-augmented generation (RAG), global planning across the whole system, refinement of the policies that govern lower-tier agents, and access to external tools and APIs. The simulation sandbox and the digital twin also allow candidate plans to be tested before they are released to the physical system.

The tiers are linked by two vertical information flows moving in opposite directions. Data, states, feedback, events, logs, and traces flow upward, giving higher tiers a progressively wider view of the system. Goals, policies, plans, and commands flow downward, translating high-level objectives into increasingly concrete instructions. Alongside these vertical flows, each tier supports horizontal agent-to-agent collaboration, so peer agents can share tasks and state without routing every interaction through a higher layer.

What distinguishes this design from conventional layered IoT architectures is the \textit{cross-layer agentic intelligence plane} shown on the right of Fig.~\ref{Aiot_architecture}. The plane provides two elements that all agents share regardless of where they run. The first is the agentic loop (perceive, reason, plan, act, learn), which defines a common behavioral model for every agent in the system and is examined in Section~\ref{sec_III}B. The second is a set of shared agentic modules for memory, tool use, communication, coordination, and safety and governance, implemented once and reused across tiers rather than re-created inside each layer. Placing safety and governance in this shared plane subjects every autonomous action to the same constraints, whether it is a single actuator command at the device level or a global policy update in the cloud. Intelligence, in this architecture, is not the property of one ``smart'' layer; it is a system-wide capability scaled to the resources of each tier.

\subsection{Agent Cognitive Loop}
At the core of Agentic IoT lies a structure that enables agents to move beyond simple rule execution and presents the processes of sensing, reasoning, planning, acting, and learning within a holistic cognitive loop, shown as the Agentic Loop in Fig.~\ref{Aiot_architecture}.

The loop begins with perception. The agent collects raw observations, including sensor readings, system states, events, and messages from other agents, and converts them into a structured context. In the reasoning stage, the agent interprets this context: it determines what is happening, why it is happening, and what it implies for its current goals. Unlike a threshold rule, this stage can weigh several pieces of evidence together and may employ explicit reasoning strategies such as chain-of-thought or ReAct (Table~\ref{tab:comparison}). Planning then turns conclusions into a course of action. The agent decomposes its goal into steps, selects a tool, an actuator, or a collaborating agent for each step, and revises the plan when conditions change. In the acting stage, the plan is executed: devices are actuated, APIs and tools are called, and subtasks are delegated to other agents. Learning closes the cycle. The agent compares outcomes with its expectations, stores the experience in memory, and adjusts its future reasoning and planning accordingly.

Two properties separate this loop from the control loops of traditional IoT systems. First, it is stateful: each iteration is informed by the memory of earlier ones, so behavior can improve over time instead of repeating a fixed response. Second, the loop runs at a different depth and timescale in each tier. A device agent completes fast, shallow iterations within milliseconds; an edge agent iterates over seconds while coordinating local tasks; a cloud agent may take minutes or hours for global planning and policy refinement. Because every tier implements the same loop, these cycles nest naturally: the outcome of one slow cloud-level iteration becomes the set of goals and policies within which the many faster iterations below it operate.

\subsection{Memory, RAG and Tool Integration}
A cognitive loop is only as capable as the state it can keep and the actions it can take. In the proposed architecture, both concerns are handled by shared modules of the agentic intelligence plane (Fig.~\ref{Aiot_architecture}) rather than being left as internal details of individual models.

Memory is organized in two forms that mirror the tier structure. Short-term memory, held by edge and device agents, stores the immediate operational context: recent observations, the state of the task in progress, and the latest instructions received from above. Long-term memory, maintained in the cloud, accumulates episodic and semantic knowledge over the system's lifetime: past incidents and their resolutions, device behavior profiles, seasonal patterns, and refined policies. Together, they allow an agent to interpret a situation not only through what it currently observes but also through what the system has observed before.

RAG connects this long-term memory to the reasoning stage. Instead of expecting a language model to carry every relevant fact in its parameters, the agent retrieves the pertinent knowledge at decision time (device manuals, historical logs, maintenance records, domain documents) and places it in the model's context. This grounds decisions in verifiable, system-specific sources, reduces the risk of hallucinated outputs, keeps knowledge updatable without retraining, and lets even the small models of the lower tiers act with information well beyond their parameter capacity.

Tool integration plays the complementary role: where memory and retrieval extend what an agent knows, tools extend what it can do. Through function calling, an agent invokes external capabilities as steps of its plan, such as querying a database, requesting a forecast, running a what-if scenario in the simulation sandbox, or issuing a command through a device API. Tool use in Agentic IoT, however, has a property that purely digital agent systems lack: some tool calls produce physical effects. For this reason, tool invocations pass through the shared safety and governance module, which validates them against operational constraints before execution.
\begin{table*}[t]
\caption{(Small) Language Models for Agentic IoT: A Comparative Overview}
\label{tab:iot_llm_comparison}
\centering
\setlength{\tabcolsep}{4pt}
\renewcommand{\arraystretch}{1.15}
\begin{tabular}{p{2.4cm}p{1.2cm}p{2.4cm}p{1.6cm}p{3.5cm}p{0.9cm}p{1.6cm}p{1.5cm}}
\hline\hline
\textbf{Model} & \textbf{Params} & \textbf{Architecture} & \textbf{Deployment} & \textbf{Capability} & \textbf{Tool} & \textbf{Multimodal} & \textbf{Quantization} \\
\hline
DeepSeek-R1 Distill & 1.5B / 7B / 8B & Transformer & Edge / GPU & Chain-of-thought reasoning & Yes & Text & INT4, INT8 \\
Falcon 3 & 3B / 7B & Transformer (GQA) & Edge / GPU & General NLP, reasoning & Yes & Text & INT4, INT8 \\
Falcon-E & 1B / 3B & Hybrid (Trans\-former + Mamba) & MCU / Edge & CPU-only inference, low-power NLP & Ltd. & Text & INT4, INT8 \\
Gemma 2B & 2B & Transformer & Edge & General NLP & No & Text & INT4, INT8 \\
Gemma 4 E2B & $\sim$2B (eff.) & Transformer (PLE) & Edge / Mobile & Multimodal NLP, function calling & Yes & Text, Image & INT4, INT8 \\
LLaMA 3.2 & 1B / 3B & Transformer & Edge & General NLP, tool use & Yes & Text & INT4, INT8 \\
Mistral 7B & 7B & Transformer (GQA+SWA) & GPU & General NLP & Yes & Text & INT4, INT8 \\
MobileLLaMA & 1.4B & Transformer (LLaMA) & Edge / Mobile & On-device NLP inference & Ltd. & Text & INT4, INT8 \\
Phi-2 & 2.7B & Transformer & Edge & NLP, reasoning & No & Text & INT4, INT8 \\
Phi-3 Mini & 3.8B & Transformer (dense) & Edge & Reasoning, coding & Yes & Text & INT4, INT8, FP16 \\
Phi-3.5 Mini & 3.8B & Transformer (dense) & Edge / Mobile & Multilingual NLP, long context (128K) & Yes & Text & INT4, INT8, FP16 \\
Qwen2.5 & 1.5B / 3B & Transformer & Edge & NLP, reasoning & Yes & Text & INT4, INT8 \\
Qwen2.5-Coder & 1.5B / 7B & Transformer & Edge / GPU & Code generation, tool tasks & Yes & Text & INT4, INT8 \\
Qwen2.5-Omni & 3B / 7B & Transformer + Audio/Vision Enc. & Edge / GPU & Cross-modal reasoning and tool use & Yes & Text, Image, Audio, Video & INT4, INT8 \\
SmolLM2 & 135M / 360M / 1.7B & Transformer (LLaMA) & MCU / Edge & Lightweight NLP; func. calling (1.7B only) & Ltd.$^\dagger$ & Text & INT4, INT8 \\
SmolVLM2 & 256M / 500M & Vision-Language (VLM) & MCU / Edge & Compact vision-language inference & Ltd. & Text, Image, Video & INT4, INT8 \\
TinyLlama & 1.1B & Transformer (LLaMA-2) & Edge & General NLP & No & Text & INT4, INT8 \\
\hline\hline
\multicolumn{8}{l}{\footnotesize $^\dagger$ Tool use supported only in the 1.7B-Instruct variant. Models ordered alphabetically; variants by ascending parameter count.} 
%\multicolumn{8}{l}{\footnotesize Models ordered alphabetically; variants by ascending parameter count.}
\end{tabular}
\end{table*}

\subsection{Communication: MCP, A2A, and IoT Protocol Bridging}
Communication in Agentic IoT spans two protocol families that evolved independently. Established IoT protocols such as MQTT, CoAP, REST, and WebSocket were designed for constrained devices; they carry telemetry and commands efficiently but encode no semantics for goals, capabilities, or negotiation. Emerging agent protocols supply exactly this missing layer \cite{ehtesham2025survey}. The Model Context Protocol (MCP) standardizes how an agent connects to external tools, data sources, and contextual resources through a uniform interface. The Agent-to-Agent (A2A) protocol standardizes how agents discover one another, advertise their capabilities, delegate tasks, and exchange results, and protocol suites such as ACPs extend this direction with dedicated mechanisms for agent registration, discovery, interaction, and collaboration~\cite{li2025acps}. In the reference architecture, these agent-level protocols operate between the cloud and edge tiers, while device-level protocols connect the edge and device tiers (Fig.~\ref{Aiot_architecture}).

Neither family can replace the other. Resource-constrained devices cannot host agent protocol stacks, and agent protocols are not built for high-frequency telemetry. The architecture therefore assigns protocol bridging to the edge tier, where edge agents translate between the two worlds in both directions. Upward, they convert raw device traffic, such as an MQTT sensor message or a CoAP response, into structured observations that agents can reason over. Downward, they turn agent decisions into concrete commands issued over whatever protocol each device understands. The bridge is more than format conversion: it aggregates, filters, and enriches information, forming the semantic boundary at which data becomes context and plans become commands. It also decouples the two sides, so the agentic layer can adopt new protocols and models without modifying the installed device base.

\subsection{Deployment Strategies: Device, Edge/Fog, Cloud, Hybrid}
As shown in Fig.~\ref{Aiot_architecture}, deployment in Agentic IoT is considered through the traditional three-layer IoT architecture. A device-level deployment approach is presented at the physical/device layer, an edge/fog deployment approach at the intermediate layer, and a cloud deployment approach at the intelligence layer. The layer in which an agent will be implemented and executed can be determined by considering the requirements of the application domain, such as decision scope, response time, and the amount of data to be transferred. Cloud deployment offers the most comprehensive reasoning capacity due to the use of large models, long-term memory, and system-wide visibility. This deployment is suitable for strategic and latency-tolerant tasks such as optimization, policy generation, and cross-site analytics.

In edge/fog deployment, priorities change. Quantized medium-sized language models (Table~\ref{tab:iot_llm_comparison}) running on gateways and edge servers can respond within local latency limits, keep data local, and continue operating when the uplink is interrupted. However, this approach has to operate with more limited model capacity, memory, and context length. Therefore, edge placement becomes more suitable for locally scoped, repetitive, time-critical, or privacy-sensitive decisions rather than open-ended and system-wide reasoning. In on-device deployment, on-premises solutions can be provided by considering data privacy and security, as well as fast response time, mostly for small language models.

On the other hand, instead of using each layer alone for deployment, we present an architecture that combines them in a hybrid manner. Here, we position hybrid deployment not as a simple compromise between the cloud and the device, but as the fundamental operating model for Agentic IoT systems. In this structure, responsibilities are distributed across layers. In this deployment, device agents perform immediate perception-action behaviors, edge agents make tactical decisions in line with the policies received from upper layers, and cloud agents carry out strategic planning and long-term learning processes. The transfer of tasks to upper layers is performed according to the requirements of the relevant task. Therefore, while routine situations can be resolved locally, novel or high-risk situations are transferred to upper layers because they require more comprehensive reasoning. In this way, more comprehensive reasoning is performed by tolerating higher latency. In cases where connectivity is interrupted, edge and device agents continue to operate according to the most recently delivered policies, and the system experiences gradual performance degradation instead of stopping completely. This hierarchical distribution of intelligence enables Agentic IoT systems to respond quickly on the device and edge side, while also benefiting from deeper cloud-scale reasoning capacity.

\section{Domain-Specific Agentic IoT Applications}
\label{sec_IV}
This section reviews existing agentic IoT systems and approaches proposed in recent studies. These works focus on enabling autonomous behavior, coordination, and intelligent decision making through various frameworks and models. The approaches are analyzed in terms of their methodology, architecture and key results.
\subsection{IoT \& Agent Infrastructure}
In this subsection, we examine studies focusing on potential agent architectures and deployment infrastructures within the scope of agentic IoT.

Yu et al.\cite{yu2013internet} propose an Internet of Agents paradigm to address the limited adoption of multi-agent systems in IoT. They highlight the complexity of existing approaches and emphasize the need for accessible tools for end-users. They review existing agent-oriented models and tools such as JADE, the BDI model, Prometheus, Tropos, and Goal Net, showing that they are mainly designed for expert developers.The approach is based on intelligent agents, emphasizing reasoning, autonomy, and user-driven customization. The study concludes that simplifying these models and enabling end-user participation is essential for wider adoption of agent-based IoT systems.
Pico-Valencia et al. \cite{pico2016semantic} propose a semantic agent contract model for IoA to improve interoperability among agents. They use an OWL-based ontology (IoA-OWL) with linked data to describe agents, services, and context. The model defines standardized semantic profiles, including agent, context, service, social, model, and object layers, to represent agent capabilities and relationships. The approach is based on semantic (ontology-driven) agents, enabling structured knowledge representation and consistent interaction across heterogeneous systems. The study concludes that this model enhances interoperability and coordination in IoA environments.Pico-Valencia et al. \cite{pico2018integration1} propose a MAS-ROA architecture to integrate multi-agent systems with IoT for managing smart objects. They use RESTful resource-oriented services and workflow-driven agents to enable adaptive and collaborative control. The architecture follows a multi-layer design where agents interact with IoT resources through RESTful services and dynamic workflows. The approach is based on reactive agents within a multi-agent system, where agent behavior is driven by workflow-based control and stimulus–response mechanisms. The study shows that this model improves adaptability, interoperability, and collaborative control in IoT environments.
Pico-Valencia et al.\cite{pico2018towards} analyze the IoT paradigm from an intelligence and autonomy perspective, proposing the Internet of Agents as an evolution to enhance these capabilities. They examine multiple application domains and present IoA-based adaptations of existing IoT systems. The study analyzes 24 IoT application scenarios and redesigns them using agent-based approaches across domains such as smart industry, smart city, and healthcare. The approach is based on intelligent software agents. It focuses on autonomy, learning, and collaboration to improve system behavior. The study concludes that adding these agents to IoT environments can significantly improve intelligence, adaptability, and interoperability.
Pico-Valencia et al.\cite{pico2019systematic} propose a systematic method for building IoA systems from existing IoT infrastructures. They use a bottom-up and agile-based approach. In this approach, IoT objects are modeled as agents called LOAs. The method also uses semantic contracts and workflows. It is organized into iterative phases. These phases help transform IoT components into agent-based systems and support continuous improvement. The approach is based on semantic, ontology-driven agents. LOAs interact by using structured knowledge and workflow coordination. The study shows that this method improves scalability, adaptability, and interoperability in IoA system development.
Rodriguez-Benitez et al. \cite{rodriguez2022internet} propose an IoA-based architecture for distributed training and deployment of CNN models in IoT environments. They use a FIPA-compliant multi-agent system with specialized agents (e.g., Trainer, Deployment, Monitor) and communication via MQTT and FIPA-ACL. The system applies model quantization and TCP-based transmission to enable execution on resource-constrained devices, achieving efficient and accurate deployment.

\subsection{Security \& Trust}
In this subsection, we examine studies focusing on threat detection, trust establishment, and secure coordination.

Aref et al. \cite{aref2017acting} propose a trust establishment model (ATeIF) for IoA that enables agents to adapt their behavior using implicit feedback. They use retention-based metrics and reinforcement learning concepts to estimate trust without explicit user input. The study models trust using multi-criteria service evaluation and retention-based learning to adjust agent behavior dynamically. The approach is based on intelligent agents within a multi-agent system (MAS). The study concludes that the proposed model improves trustworthiness and increases successful interactions among agents.
Aref et al. \cite{aref2020integrated} propose an integrated trust establishment model for IoA that combines multiple feedback mechanisms to improve agent behavior. They use both direct and indirect trust signals. These signals help adjust agent interactions dynamically. The study combines reputation, experience, and contextual factors to calculate trust values in a more complete way. The approach is based on intelligent agents in a multi-agent system. The study concludes that the proposed model improves trust accuracy and supports more reliable interactions among agents. Li et al. \cite{li2025acps} propose a unified protocol framework called ACPs for the Internet of Agents. The aim is to solve interoperability and coordination problems. They define standard protocols for agent registration, agent discovery, agent interaction, and agent collaboration. These protocols include ARP, ADP, AIP, and ATP. The study introduces a layered IoA architecture and supports capability orchestration and task decomposition across heterogeneous agents. It also considers the integration of large language model (LLM)-based agents to enhance communication and intelligent interaction. The approach is based on intelligent, heterogeneous multi-agent systems (MAS), enabling scalable and coordinated agent interactions. The study concludes that the proposed framework improves interoperability, collaboration, and adaptability in IoA environments.
Kumia et al. \cite{kumi2025securing} propose a dual-proxy, multi-plane gateway architecture to secure agentic AI in IoT systems. They use separate planes for data, control, security, and coordination, along with a sentinel mechanism and digital twin for safe execution. The study introduces a seven-plane architecture and a goal–plan–step validation mechanism to ensure safe and verifiable agent actions. It incorporates LLM-based agents and supports interaction with models such as GPT-4, along with tool-augmented frameworks and RAG-based context integration. The approach is based on intelligent, agentic AI systems integrated with IoT environments. The study concludes that the proposed architecture improves security, governance, and reliability in agentic IoT systems.
Ren et al. \cite{ren2026toward} propose the Internet of Agentic AI (IoA) paradigm to enable scalable and intent-driven coordination among agents. They introduce a layered architecture with core protocols and an intelligence-based reinforcement learning (IRL) framework for multi-agent coordination. The study enables dynamic task composition, semantic interaction, and distributed planning among heterogeneous agents. It also incorporates cognitive capabilities supported by LLM-based agents and tool-augmented frameworks (e.g., AutoGen, CrewAI, LangGraph). The approach is based on intelligent, heterogeneous multi-agent systems (MAS) combined with reinforcement learning. The study concludes that the proposed IoA paradigm improves adaptability, coordination efficiency, and scalability.
Vijetha \cite{vijetha2026agentic} proposes a self-adaptive agentic framework called AISAF. AISAF is designed for threat detection in cloud, edge, and IoT systems. The approach uses a hybrid CNN-LSTM Transformer model. It also uses a drift-aware meta-optimization mechanism for selective adaptation. The framework detects changes in the latent space and monitors confidence scores. These mechanisms allow the system to trigger model updates automatically. They also help the system maintain performance when data distributions change over time. The framework includes an attention-based explainability layer. This layer makes the decision-making process more transparent and consistent during adaptation. The approach is based on intelligent agentic AI systems in a heterogeneous multi-agent environment. The study shows that AISAF improves adaptability, robustness, and recovery from data drift in dynamic cybersecurity environments.
\begin{table*}[t]
\renewcommand{\arraystretch}{1.3} % Satır aralığını rahatlatır
\caption{Comparative Analysis of Internet of Agents and Agentic IoT Studies Across IoT Application Domains}
\label{table:ioa_literature}
\centering
\footnotesize
\begin{tabularx}{\textwidth}{@{}l p{3.5cm} c p{2.8cm} l l p{4.0cm}@{}}
\toprule
\textbf{IoT Domain} & \textbf{Study} & \textbf{Year} & \textbf{Problem Context} & \textbf{Taxonomy} & \textbf{Architecture} & \textbf{Underlying Mechanism} \\ \midrule

\multirow{6}{*}{Infrastructure} 
 & Yu et al.~\cite{yu2013internet} & 2013 & Resource Mgmt. & Cognitive & Intelligent & BDI (Belief-Desire-Intention) \\
 & Pico-Valencia and Holgado-Terriza~\cite{pico2016semantic} & 2016 & Interoperability & Cognitive & Semantic & BDI (Belief-Desire-Intention) \\
 & Pico-Valencia and Holgado-Terriza~\cite{pico2018integration} & 2018 & System Integration & Cognitive & Reactive & N/A \\
 & Pico-Valencia et al.~\cite{pico2018towards} & 2018 & Architecture Analysis & Operational & Software & FIPA-ACL (Agent Comm. Lang.) \\
 & Pico-Valencia et al.~\cite{pico2019systematic} & 2019 & Interoperability & Cognitive & Semantic & LOA (Linked Open Agent) \\
 & Rodriguez-Benitez et al.~\cite{rodriguez2022internet} & 2022 & Deep Learning Ops & Operational & Specialized & TFLite (Tensorflow Lite) \\ \hline

\multirow{6}{*}{Security} 
 & Aref and Tran~\cite{aref2017acting} & 2017 & Trust Modeling & Distributed & Heterogeneous & Implicit Feedback \\
 & Aref and Tran~\cite{aref2020integrated} & 2020 & Trust Establishment & Cognitive & Intelligent & Q-Learning \\
 & Li et al.~\cite{li2025acps} & 2025 & Multi-Agent Collab. & Distributed & Heterogeneous & Protocol-Based \\
 & Kumi et al.~\cite{kumi2025securing} & 2025 & Governance & Autonomous & Agentic AI & LLM-based Reasoning \\
 & Ren et al.~\cite{ren2026toward} & 2026 & Protocol Design & Protocol-Driven & Layered IoA & RL (Reinforcement Learning) \\
 & Vijetha~\cite{vijetha2026agentic} & 2026 & Threat Detection & Autonomous & Agentic AI & CNN-LSTM-Transformer \\ \hline

\multirow{5}{*}{Network} 
 & Pico-Valencia et al.~\cite{pico2019towards} & 2019 & Data Interoperability & Cognitive & Intelligent & LOA (Linked Open Agent) \\
 & Jeong and Kountouris~\cite{jeong2025draco} & 2025 & Federated Learning & Distributed & Decentralized & SGD (Stoch. Grad. Descent) \\
 & Jiang et al.~\cite{jiang2025stackelberg} & 2025 & Task Offloading & Operational & Embodied AI & GT (Game-Theoretic Model) \\
 & Elewah et al.~\cite{elewah2025agentic} & 2025 & Real-time Search & Distributed & MAS & RAG (Retrieval-Augm. Gen.) \\
 & Vashisht et al.~\cite{vashisht2026multimodal} & 2026 & Network Slicing & Distributed & Supervisory & Ensemble Learning \\ \hline

\multirow{3}{*}{Smart Cities} 
 & Bui and Jung~\cite{bui2018internet} & 2018 & Traffic Control & Cognitive & Intelligent & GME (Group Mutual Exclusion) \\
 & Kamel~\cite{kamel2025conceptual} & 2025 & Disaster Response & Distributed & Heterogeneous & YOLOv8 (Object Detection) \\
 & Chumyen~\cite{chumyen2025multi} & 2025 & Water Distribution & Distributed & Decentralized & PPO (Prox. Policy Optim.) \\ \hline

\multirow{5}{*}{Industry} 
 & Tariq et al.~\cite{tariq2025edge} & 2025 & Precision Farming & Cognitive & Rule-based & MiT-B0 (Vision Transformer) \\
 & Kandamali et al.~\cite{kandamali2025cottonbot} & 2025 & Irrigation Advisory & Autonomous & Tool-Use LLM & LLM (Llama 3.1) \\
 & Sümer et al.~\cite{sumer2025smart} & 2025 & Machine Customers & Semi-autonomous & Blockchain-IoT & RBS (Rule-Based System) \\
 & Petrović et al.~\cite{petrovic2025llm} & 2025 & Sustainability & Design-time & Autonomous & LLM (GPT-4o) \\
 & Mohanaprasad et al.~\cite{mohanaprasad2026harmony} & 2026 & Smart Home Orch. & Autonomous & Multimodal & VLM (Qwen 2.5 VL) \\ \bottomrule

\multicolumn{7}{@{}l}{\rule{0pt}{3ex}}
\end{tabularx}
\end{table*}

\subsection{Network}
In this subsection, we examine network-based studies focusing on distributed and edge-computing-based learning, as well as intelligent resource and task management.

Pico-Valencia et al. \cite{pico2019towards} analyze how IoT systems are evolving toward the Internet of Agents paradigm. The goal is to make IoT systems more autonomous and intelligent. They examine the limitations of traditional IoT architectures. They also propose agent-based adaptations to support more flexible and intelligent system behavior. The study presents a layered IoA architecture. It also identifies the main components needed for agent integration. These components include communication, coordination, and semantic interaction mechanisms. The study explains how agent-based transformation can improve system responsiveness and decision-making. The approach is based on intelligent agents in a multi-agent system. The study concludes that IoA can significantly improve adaptability, interoperability, and autonomy in IoT environments.
Jeong et al. \cite{jeong2025draco} propose DRACO, a decentralized and asynchronous federated learning framework for distributed AI in IoT environments. They design an agent-based system that allows devices to train models together without central coordination. The system supports asynchronous updates and peer to peer communication. The framework includes mechanisms for dynamic participation, model aggregation, and communication efficiency under heterogeneous conditions. It also improves scalability and robustness by reducing dependence on central servers and handling device differences. The approach is based on intelligent agents in a decentralized multi-agent setting. The study concludes that DRACO improves scalability, efficiency, and resilience in distributed IoT learning systems.
Jiang et al. \cite{jiang2025stackelberg} propose a Stackelberg game-based framework for agentic AI task offloading in the Internet of Agents. The goal is to address resource limitations. They model interactions among wireless, mobile, fixed, and aerial agents. This allows hierarchical and efficient task delegation. The study formulates the problem as a leader–follower game to optimize pricing and offloading strategies under heterogeneous conditions. It considers agentic AI tasks supported by large models such as LLMs and VLMs for reasoning and decision-making. The approach is based on intelligent, heterogeneous multi-agent systems. The study concludes that the method improves resource efficiency and task execution performance.
Elewah et al. \cite{elewah2025agentic} propose an agentic search engine framework for real-time information retrieval in dynamic environments. They design an agent-based system that enables autonomous query understanding, task decomposition, and multi-source information aggregation. The framework supports real-time decision-making and adaptive search through coordinated agent interactions and context-aware processing. It incorporates LLM-based components to enhance natural language understanding and reasoning during query execution. The approach is based on intelligent and heterogeneous multi agent systems. The study concludes that the proposed system improves search accuracy, responsiveness, and adaptability in real time scenarios.
Vashisht et al. \cite{vashisht2026multimodal} propose an agentic intelligence-assisted machine learning fusion framework for sustainable 6G network slicing. The framework is designed for dynamic and multimodal conditions. They use a stacked ensemble model that combines LightGBM, Random Forest, and Logistic Regression. They also use an agentic overlay to enforce SLA constraints in real time. The framework combines multimodal features and supports adaptive decision-making through a hybrid machine learning and agentic architecture. It allows slicing decisions to be checked and adjusted according to SLA requirements. These requirements include throughput, latency, and energy consumption. The approach is based on intelligent agents in a multi agent system. The study concludes that the proposed framework improves accuracy, stability, and SLA compliance in 6G environments.

\subsection{Smart Cities \& Disaster}
In this subsection, we present studies that address issues such as emergency response, disaster management, and mobility in smart cities within the IoA framework.

Bui et al. \cite{bui2018internet} propose an Internet of Agents framework for connected vehicles. The goal is to support distributed traffic control without using centralized infrastructure. In this framework, each vehicle is modeled as an agent. Vehicles communicate with each other through vehicle to vehicle communication. They coordinate their decisions by using an extended Ricart Agrawala algorithm based on group mutual exclusion. The approach is based on intelligent agents in a multi-agent system. The study concludes that the proposed framework improves traffic efficiency and reduces waiting time compared to traditional traffic systems.
Kamel et al. \cite{kamel2025conceptual} propose an integrated IoA - IoT framework called AIMS-AI for smart disaster detection and response. They design a multi layered architecture for real time sensing and coordinated decision making among heterogeneous agents. The framework uses AI models such as YOLOv8, BERT, and large language models to improve perception and response. The approach is based on intelligent and heterogeneous multi-agent systems. The study concludes that AIMS-AI improves scalability and real-time response performance.
Chumyen et al. \cite{chumyen2025multi} propose a multi-agent IoT-based smart water distribution system for disaster relief in smart cities. They design autonomous dispenser agents integrating computer vision (MobileNet-SSD) and LSTM-based demand forecasting to enable real-time perception and prediction. The system uses multi-agent reinforcement learning (PPO) for decentralized coordination under resource-constrained environments. The approach is based on intelligent, heterogeneous multi-agent systems. The study concludes that it improves fairness, efficiency, and resilience in emergency water distribution.

\subsection{Industry \& Farming}
In this subsection, studies on the industrial, agricultural, and smart-environment applications of agentic IoT are presented.

Tariq et al. \cite{tariq2025edge} propose an agent-based IoT framework to enable intelligent coordination and decision-making in distributed environments. They design a layered architecture that integrates sensing, communication, and agent-based processing to support adaptive system behavior. The study enables decentralized interaction and dynamic task execution among agents under heterogeneous conditions. The approach is based on intelligent agents within a multi-agent system (MAS). The study concludes that the proposed framework improves system efficiency, scalability, and coordination.
Kandamali et al. \cite{kandamali2025cottonbot} propose an AI-driven agricultural assistant (CottonBot) that integrates RAG and agentic AI tools to support real-time decision-making in cotton farming. They design a hybrid architecture combining domain-specific knowledge retrieval with IoT sensor data and weather APIs to generate context-aware and field-specific recommendations. The system enables dynamic irrigation decisions through agent-based tool calling and real-time data processing. The approach is based on intelligent agents within a multi-agent system (MAS). The study utilizes open-weight LLMs, particularly LLaMA 3.1 via the Ollama framework, along with models such as Mistral and Phi-3 for evaluation.
Sümer et al. \cite{sumer2025smart} propose a semi-autonomous IoT framework introducing the concept of “smart agents as customers” for Industry 5.0 environments. They design a system integrating IoT devices, blockchain, and smart contracts to enable automated purchasing and supply management with human-in-the-loop control. The framework supports decentralized transactions and transparent tracking of interactions through Web 3.0 technologies. The approach is based on intelligent, heterogeneous multi-agent systems. The study concludes that it improves automation, transparency, and efficiency in digital commerce and smart environments.
Petrović et al. \cite{petrovic2025llm} propose an LLM-driven approach to automate scalability and maintenance of IoT systems using model-driven engineering (MDE). They use system modeling, constraint validation (OCL), and automated code generation to support hardware integration and maintenance tasks. The framework enables automated system updates and decision-making through LLM-based reasoning and planning agents. It incorporates models such as GPT-4o and o1-mini for code generation and maintenance planning. The approach is based on intelligent, heterogeneous agent-based systems. The study concludes that it reduces cognitive load and improves efficiency in IoT system management.
Mohanaprasad et al. \cite{mohanaprasad2026harmony} propose HARMONY, a multimodal LLM-driven framework for smart home IoT systems to enable context-aware and intent-driven automation. They design a multi-layered architecture where an MLLM core performs multimodal perception, intent recognition, and task planning, while an orchestrator coordinates domain-specific agents. The framework uses multimodal large language models for reasoning and orchestration. It includes models such as Qwen 2.5 VL, GPT-4.1, Gemini 2.5, and LLaMA. The approach is based on intelligent and heterogeneous multi agent systems. The study concludes that this approach improves personalization, coordination, and decision making in smart home environments.

\section{Challenges and Future Research Directions}
\label{sec_V}
Agentic IoT has strong potential to transform traditional networks into autonomous, adaptive, and intelligent ecosystems. However, it also creates several important challenges. These challenges must be solved before agentic IoT can be deployed reliably and used in real world systems.

\begin{figure*}[!t]
    \centering
\includegraphics[width=\textwidth]{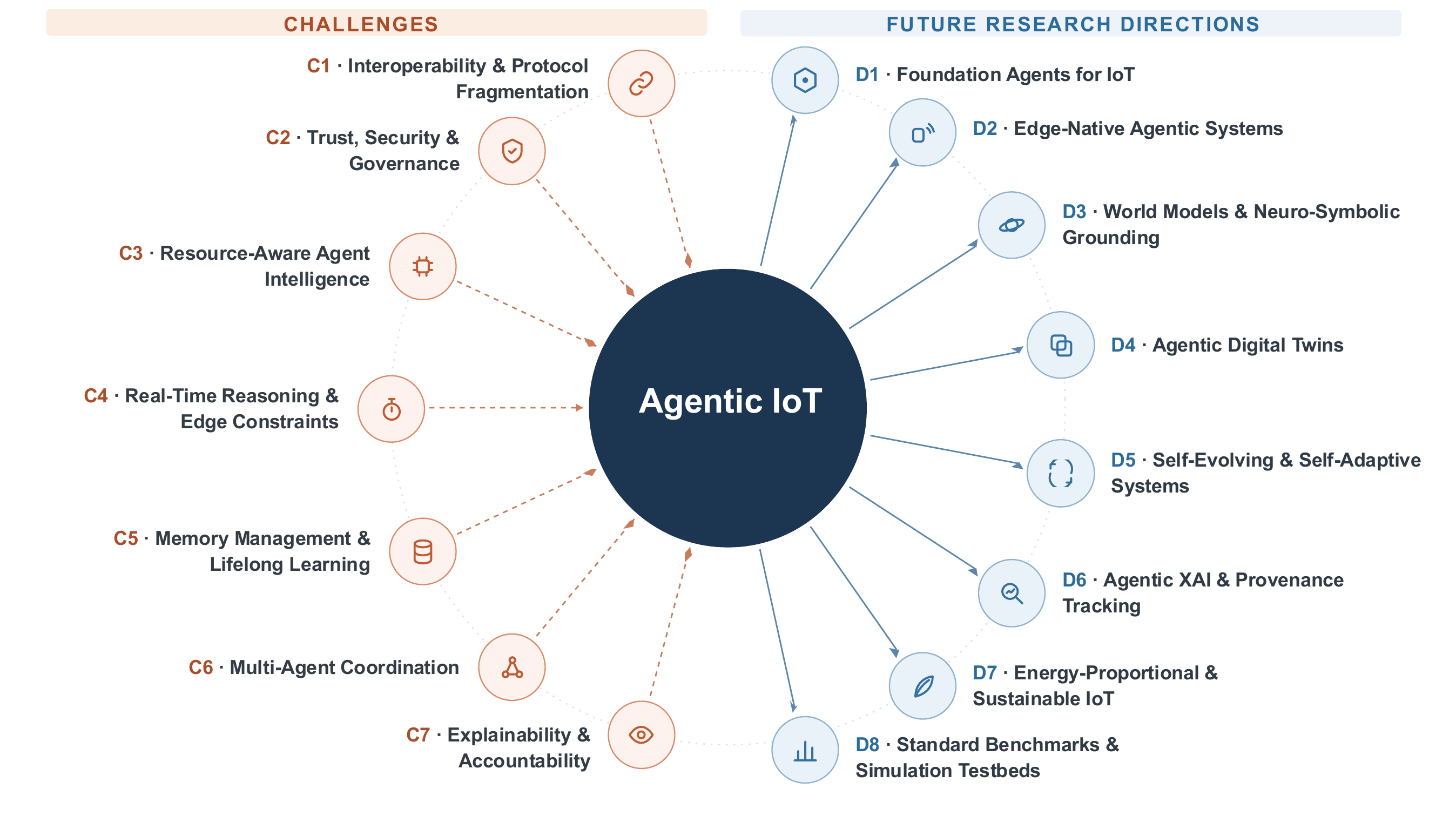}
    \caption{Challenges and future research directions in Agentic IoT.}
    \label{challenges_FRD}
\end{figure*}

\subsection{Challenges}

\subsubsection{Interoperability, Semantic Standardization, and Protocol Fragmentation}
Interoperability, semantic standardization, and protocol fragmentation are major challenges for Agentic IoT systems. An Agentic IoT system must allow agents, devices, and services from different manufacturers to work together in the same environment. However, these components often use different data formats, APIs, and communication rules. These differences make many components incompatible with each other. As a result, each integration often becomes manual, fragile, and difficult to reuse. This also makes the system harder to scale, secure, and maintain \cite{ehtesham2025survey}. New protocols such as MCP, A2A, and ACP have begun to offer standardized frameworks for tool access, inter-agent messaging, capability discovery, and task sharing. However, these protocols are still new and at an early stage of maturity, and they are not fully integrated with established IoT communication infrastructures~\cite{li2025acps}. In this context, the main challenge is to create a common semantic and protocol layer in which heterogeneous agents and resource-constrained devices can discover each other, declare their capabilities in a machine-readable way, share semantically consistent data, and work together without requiring a custom middleware layer for each use case.

\subsubsection{Trust, Security, and Governance}
When an IoT agent not only collects data but also makes decisions and takes actions on its own, the threat space goes far beyond data privacy. New attack paths come to the fore, such as steering the agent’s reasoning, pushing it toward unsafe tool calls, or taking over inter-agent collaboration \cite{11311635}. Existing IoT security methods, however, are not designed for systems that can plan and act on their own; moreover, agent chains composed of multiple layers and manufacturers make authentication, tracing the source of a decision, and determining responsibility even more difficult \cite{liu2025secure}. Especially in safety-critical applications, how these systems will be audited remains unanswered, starting with the question of who is responsible when an autonomous decision causes harm.
\subsubsection{Resource-Aware Agent Intelligence}
Capabilities such as reasoning, planning, memory, and tool use, which make agents intelligent, are mostly fed by large language models or large-scale foundation models. However, the computing power, memory, bandwidth, and energy requirements of these models exceed the capacity of most IoT devices \cite{belcak2025small}. Although small language models are becoming increasingly powerful, adjusting model capacity according to the difficulty of the task, the risk level of the decision, and the limits of the hardware remains a challenging task \cite{dikici2026small}. The main challenge here is to design the agent from the beginning in a resource-aware, scalable, and task-sensitive manner without compromising features such as accuracy, security, context awareness, and explainability. In such an approach, the same agent architecture should be able to operate as an elastic intelligence layer that can gradually decrease and increase its capability from powerful cloud servers to edge gateways and then down to microcontroller-level devices.
\subsubsection{Real-Time Reasoning and Edge Constraints}
Multi-step reasoning, information retrieval, memory access, tool calling, and inter-agent coordination processes naturally create additional latency. However, in many IoT system loops, the decision must be made within milliseconds \cite{park2026minimizing}. Moving reasoning to the edge device reduces the round-trip time to the server, but this time the agent has to cope with limited computing power, disconnected connections, and often incomplete or uncertain observations \cite{wang2025cognitive}. Therefore, the main problem is not only to run a faster model, but also to establish a risk-sensitive and latency-aware reasoning architecture that dynamically determines which decisions will be made on the local server or in the cloud.
\subsubsection{Memory Management and Lifelong Learning}
For an agent to perform its defined operational task well, it also needs to remember the context, past interactions, the outcomes of previous decisions, and the information it has learned from the environment. However, due to reasons such as a limited context window, narrow device storage, and energy constraints, an IoT agent cannot effectively decide how much of this information it can reliably store and retrieve \cite{xing2020reliability}. Therefore, the challenge is not merely to add a larger memory to the system, but to develop reliable memory management mechanisms that can decide which information will be stored, when it will be updated, when it will be forgotten, and which information will be retrieved at the moment of decision.
\subsubsection{Multi-Agent Coordination and Collective Intelligence}
In Agentic IoT, many different agents need to communicate with one another, share tasks, coordinate resources, and resolve conflicts under changing conditions and with limited information. This requirement creates the necessity for tasks to be performed by multiple agents \cite{tran2025multi}. At this point, as the number of agents increases, the communication load, coordination complexity, and risk of unexpected collective behavior also increase. Therefore, the goal in multi-agent systems is not only for each agent to perform its own local tasks in the best way, but also to produce a consistent and safe collective behavior aligned with the overall goals of the system \cite{iotllm}. The challenge here is to transform agents into true collective intelligence without collectively dragging the system into instability, resource competition, or unexpected errors.
\subsubsection{Explainability and Responsible Agentic IoT}
Since agents act through a multi-step process rather than producing a single output, explaining why an IoT agent perceived, planned, and behaved in a certain way is much more difficult than interpreting the result of a fixed model\cite{zhu2026interpreting} . In IoT environments that produce physical consequences, this uncertainty weakens not only user trust but also debugging, certification, regulatory compliance, and legal accountability. Therefore, in Agentic IoT, explainability should go beyond classical XAI methods and produce process-based explanations that cover the entire decision-making process, sensor and data provenance tracking, tool-call logs, inter-agent interaction history, and human-auditable decision traces.

\subsection{Future Research Directions}
The emergence of Agentic IoT brings with it a broad and exciting research agenda that goes far beyond integrating intelligent agents with connected devices. Future studies need to focus on how these systems can reason under physical constraints, operate efficiently on edge devices, adapt over time, explain their decisions, and be evaluated through realistic benchmark tests. In this direction, the following topics reveal the main challenges in the maturation process of Agentic IoT and the opportunities ahead.

\subsubsection{Foundation Agents for IoT}
Agent research is converging on the concept of a \textit{foundation agent}, whose cognitive core is a foundation model and which is enhanced with modular memory, planning, tool use, and perception \cite{xu2025deploying}. However, these general-purpose agents have not been grounded in the physical constraints of IoT, while IoT-oriented foundation models have remained narrow and task-specific perception backbones lacking autonomous reasoning \cite{wei2025survey}. Therefore, future studies should develop domain-specific foundation agents that integrate a general reasoning core with SLM models and local perception backbones suitable for the nature of IoT; thus, a single agent can generalize across different devices and tasks with minimal labeled data and on-device efficiency.
\subsubsection{Edge-Native Agentic Systems}
Studies on Edge General Intelligence propose running agentic AI directly on edge devices and distributing capabilities such as perception, reasoning, and planning across the device–edge–cloud continuum \cite{zhang2026toward}. Most existing agents, however, still assume cloud-scale computing power and uninterrupted connectivity. Therefore, current systems are not compatible with the latency, energy, and intermittent connectivity realities of IoT. Accordingly, future studies need to develop edge-native agent designs that combine lightweight and quantized agent models with energy-aware planning. In this way, the perception–reasoning–action loop can operate on the device itself even under strict resource constraints.

\subsubsection{World Models and Neuro-Symbolic Physical Grounding}
One of the strongest research directions in Agentic IoT systems is to move the agent beyond a structure that merely reacts to current sensor data and support it with a predictive world model on which it can reason, plan, and test possible outcomes in advance \cite{jiang2026large}. Indeed, it is clear that general-purpose agents need world models that represent the causal and dynamic structure of their environments in order to make safe and consistent decisions. The literature on physical simulators, embedded intelligence, and embodied intelligence also shows that such models play a critical role in reducing the sim-to-real gap, generating counterfactual scenarios, and supporting safe planning processes \cite{ding2025understanding}. In the context of Agentic IoT, this approach refers to a transition toward explainable and autonomous IoT architectures that integrate the sensor-motor loop with a neuro-symbolic reasoning core and reduce the risk of hallucination by constraining the agent’s actions with physical laws, system constraints, and symbolic rules.
\subsubsection{Agentic Digital Twins}
Digital twins are evolving from passive copies that merely reflect the system into “agentic digital twin” structures that actively guide and reshape it \cite{burr2026agentic}. On the IoT side, twins have so far mostly remained limited to monitoring and “what-if” scenarios, and they have not had the ability to act on their own or update their models autonomously. However, when an autonomous agent is embedded into the twin, the twin can continuously monitor its physical counterpart, intervene in it, and complete the perception–action loop at its own level\cite{11568888}. As the twin’s power to affect the system it models increases, mechanisms will be needed to control risks such as performative lock-in and self-confirming model drift.
\subsubsection{Self-Evolving and Self-Adaptive Systems}
A rapidly growing research area is self-evolving agents that, instead of remaining unchanged after deployment, can improve their model, memory, tools, and even architecture by learning from experience \cite{zhao2026agentification}. Today’s Agentic IoT systems, however, are mostly based on static models or models that are retrained at certain intervals. The performance of these models decreases as conditions and sensor data change. Future research should aim to enable agents to improve themselves throughout their entire operational lifetime by bringing together lifelong learning, retrieval-augmented memory, and online adaptation. This autonomy, however, needs to be balanced with appropriate safeguards against the risk of misevolution.
\subsubsection{Agentic Explainable AI and Provenance Tracking}
Since an agent acts through a multi-step process involving tool use, memory, and coordination rather than producing a single output, what needs to be explained is no longer only a single result but the entire process itself. In Agentic IoT, where agents take physical actions that produce real-world consequences, reasoning that cannot be understood damages both trust and accountability. XAI methods in the literature, however, have mostly been designed for fixed perception models \cite{10158334}. Therefore, future studies should develop agentic XAI models adapted to IoT. Ensuring accountability at every step of the process, providing provenance information that shows which sensor a decision is based on, and creating records through which humans can trace why an agent perceived and acted in a certain way are becoming highly important for explainability.

\subsubsection{Energy-Proportional and Sustainable Agentic IoT}
The real-world scalability of Agentic IoT systems requires energy to be treated not merely as a secondary optimization metric, but as a central design constraint. Especially for agents operating on resource-constrained edge devices, quantization, pruning, early-exit mechanisms, hardware-aware inference, and networking-aware planning approaches are critically important for sustainably balancing accuracy, latency, computational cost, and energy consumption \cite{lee2026toward}. In this context, the future direction is evolving toward energy-proportional agent architectures that dynamically adjust their workload according to the complexity of the physical task, network conditions, and the risk level of the decision, rather than agents that operate continuously at fixed capacity. In such an approach, the main goal is not only to produce more accurate decisions, but also to make Agentic IoT systems long-lived and low-carbon-footprint systems by reducing the energy cost \textit{per decision}.

\subsubsection{Standard Benchmarks and Simulation Testbeds}
For the field to mature, common testbeds are needed that evaluate agentic IoT systems in a realistic and reproducible way. Recent taxonomies on agent evaluation are moving toward measuring trajectory-level behavior, reliability, and safety \cite{ferrag2026llm}. Future studies should complement these with high-fidelity simulation environments and operational metrics that enrich them along the real-time, energy, and physical-consequence dimensions of IoT. Such an infrastructure would enable different approaches to be compared on a common ground and help avoid the \textit{illusion of progress.}
\section{Conclusion}
In this paper, we examined the emerging Agentic IoT paradigm and took a first step toward establishing it as a systematic research field. We formally defined the concept, positioned it in relation to IoT, AIoT, and the Internet of Agents, and traced the paradigm shift from rule-based and model-based systems toward autonomous, agent-centric ecosystems. We then introduced a three-tier reference architecture with a cross-layer agentic intelligence plane, described the cognitive loop through which agents perceive, reason, plan, act, and learn, and explained how memory, retrieval-augmented knowledge, tool use, and protocol bridging make this loop operational across the device-edge/fog–cloud continuum. Our literature review also reveals a clear trajectory. Early Internet of Agents studies were mainly grounded in symbolic, rule-based, and ontology-driven agent architectures, whereas recent work increasingly relies on LLM-driven reasoning, agentic tool use, and dedicated agent protocols, with applications now spreading across smart cities, disaster response, network management, agriculture, and industry. The research directions presented in this paper, from foundation agents and edge-native designs to agentic digital twins and standardized benchmarks, show where progress is most needed. We hope this study provides a common ground for researchers and practitioners working on these problems and positions Agentic IoT as the practical path through which the Internet of Agents vision reaches the physical world.
\newline
\par
\textbf{Declaration on Generative AI}

During the preparation of this work, the authors used ChatGPT and Claude in order to: Grammar and spelling check. After using these services, the authors reviewed and edited the content as needed and take full responsibility for the publication’s content.
\bibliographystyle{IEEEtran}
\bibliography{references}
\end{document}